\documentclass{article}

\usepackage{microtype}
\usepackage{graphicx}
\usepackage{verbatim}
\usepackage{subfigure}
\usepackage{booktabs} 
\usepackage{caption}
\usepackage{hyperref}

\usepackage[accepted]{icml2021}

\icmltitlerunning{Hyperparameter Selection for Imitation Learning}

\begin{document}

\twocolumn[
\icmltitle{Hyperparameter Selection for Imitation Learning}



\icmlsetsymbol{equal}{*}

\begin{icmlauthorlist}
\icmlauthor{L\'eonard Hussenot}{equal,google,lille}
\icmlauthor{Marcin Andrychowicz}{equal,google}
\icmlauthor{Damien Vincent}{equal,google}
\icmlauthor{Robert Dadashi}{google}
\icmlauthor{Anton Raichuk}{google}
\icmlauthor{Lukasz Stafiniak}{google}
\icmlauthor{Sertan Girgin}{google}
\icmlauthor{Raphael Marinier}{google}
\icmlauthor{Nikola Momchev}{google}
\icmlauthor{Sabela Ramos}{google}
\icmlauthor{Manu Orsini}{google}
\icmlauthor{Olivier Bachem}{google}
\icmlauthor{Matthieu Geist}{google}
\icmlauthor{Olivier Pietquin}{google}
\end{icmlauthorlist}

\icmlaffiliation{google}{Google Research, Brain Team}
\icmlaffiliation{lille}{Univ. de Lille, CNRS, Inria Scool, UMR 9189 CRIStAL}

\icmlcorrespondingauthor{L\'eonard Hussenot}{hussenot@google.com}

\icmlkeywords{Machine Learning, Reinforcement Learning, Imitation Learning}

\vskip 0.3in
]



\printAffiliationsAndNotice{\icmlEqualContribution} 

\begin{abstract}
We address the issue of tuning hyperparameters (HPs) for imitation learning algorithms in the context of continuous-control, when the underlying reward function of the demonstrating expert cannot be observed at any time. The vast literature in imitation learning mostly considers this reward function to be available for HP selection, but this is not a realistic setting. Indeed, would this reward function be available, it could then directly be used for policy training and imitation would not be necessary.
To tackle this mostly ignored problem, we propose
a number of possible proxies to the external reward. We
evaluate them in an extensive empirical study (more than 10'000 agents across 9 environments) and make practical recommendations for selecting HPs.
Our results show that while imitation learning
algorithms are sensitive to HP choices,
it is often possible to select good enough HPs
through a proxy to the
reward function.
\end{abstract}


\section{Introduction}
Recent advances in Reinforcement Learning (RL)
now allow optimizing control policies
with respect to a given reward function even for
high-dimensional observation and action spaces \citep{dota, starcraft}.
However, in many cases it is impossible or impractical
to design a reward function which captures the desired
outcomes \citep{popov2017data}.
One of the approaches to overcome this issue
is
Imitation Learning (IL), which relies on a set
of demonstrations presenting the desired behaviour instead of
a reward signal~\citep{SCHAAL1999233,argall2009survey}. IL can be achieved through pure supervised learning~\citep{pomerleau1991efficient} but many IL approaches leverage the assumption that the expert implements an optimal policy according to an \emph{unknown} reward function. This approach, also known as Inverse Reinforcement Learning (IRL), tries to recover this unknown reward function and use an RL algorithm to train a policy to maximize it~\citep{russell1998learning,ng2000algorithms,ziebart2008maximum}. Both approaches have their advantages and drawbacks~\citep{piot2013learning}.

While all machine learning approaches require some degree of
hyperparameter (HP) tuning, the issue is especially pronounced in
RL. Indeed, RL algorithms are known to be very sensitive to the values of their numerous HPs~\citep{henderson2018deep, andrychowicz2020matters}.
RL algorithms' HPs are usually chosen by letting different agents interact with the environment and selecting the one which performed best
as measured with the environment reward.

Surprisingly, this is also a common practice in the IL domain where one does not have access to an environment reward function which accurately describes
the task.
If such a reward were known, it could be directly used to train a controller via RL.
Although expert trajectories can also be useful in this case, this is not the setting of IL, but of RL with demonstrations where \emph{both} the expert demonstrations and reward signals are used \citep{kim2013learning,piot2014boosted,hester2018deep}. This constitutes a gap between the IL framework and the experimental design of IL agents that hinders the practical utility of IL. It is indeed unclear how to tune the imitation agent without access to the reward function.

In some cases, although a per-step reward function is not available, a success signal can be computed or given by a human rater. We focus on the tasks for which (1) such signal is not available or (2) the cost of such signal is prohibitive or (3) such signal could bias the resulting policy.  Indeed, just as the reward-engineering problem leads to policies maximizing a reward in an unexpected way \citep{sims1994evolving, feldt1998generating, ecoffet2021first}, selecting policies on the basis of an incautiously designed metric or of a biased human judgement can lead to biased policies \citep{henderson2018deep}.

We present a thorough empirical study of this question
in a number of challenging domains with high dimensional spaces of actions and
observations.
We train thousands of agents, using three IL algorithms based on different paradigms, with
large parameter sweeps, and empirically compare
different HP selection strategies.
In particular, we consider a number of metrics assessing
how well the learned behaviors match the demonstrations.
Moreover, we investigate how well the algorithms perform if their
HPs are selected on a similar task
where the reward signal is assumed to be available.
As our key contributions, we:
\begin{itemize}
    \item highlight the fundamental question of HP selection in IL
    without access to an external reward signal;
    \item provide empirical evidence that the results obtained by IL algorithms
    depend heavily on how HPs are selected, confirming the prominence of the problem;
    \item propose  proxy metrics to define the task success;
    \item perform a thorough comparison of these alternatives to the
    external reward signal in a large-scale study;
    \item empirically assess the transferability of HPs across tasks for different IL algorithms;
    \item give practical recommendations for tuning HPs in IL.
\end{itemize}

\section{Hyperparameter Selection}\label{sec:hp-selection}

In this paper, we argue that HP selection should strictly follow the setup in which the algorithm is to be used. For example, when designing an offline RL algorithm, HPs should not be tuned through environment interactions
but rather offline \citep{paine2020hyperparameter}.
Similarly, in IL, one can interact with the environment but should choose HPs without access to the reward signal. We propose to choose these HPs by either (1) using proxy metrics or (2) transferring HPs from other environments that have an accessible reward.

\subsection{Using proxy metrics}
To select a model, one should use a different metric than the return
under the supposedly unknown reward.
We include a diverse set of metrics that can be used to measure the success of a policy in imitating the demonstrations. 

\textbf{Action MSE.}
The simplest way to measure similarity between agent behaviour and
a given set of demonstrations is to compare actions of the agent and the demonstrator on the states provided in the demonstrations.
In particular, for continuous control environments, we use the mean squared error (MSE) between the agent and the expert actions on the training expert states\footnote{
All action coordinates are rescaled to the $[-1, 1]$ range
to make them comparable in magnitude.}.
Notice that this can be done offline without any interaction with the environment.
For every environment, we also keep some expert trajectories as a validation set
and use it to compute the validation MSE. This will allow us to study if selecting HPs on validation data helps avoiding overfitting, notably as action MSE is actually the loss optimized by the Behavioral Cloning~\citep{pomerleau1991efficient} algorithm.

\textbf{State distribution divergence.}
Another approach to measuring how well the learned policy imitates the expert
is by comparing the distributions of states\footnote{
We only consider fully observable environments, but similar techniques
may work in the partially observable
case too.} encountered by both policies.
In particular, we compute the Wasserstein distance~\citep{villani2008optimal}
between the distribution of states in the demonstrations
and that of the agent\footnote{We approximate the Wasserstein distance using the entropy-regularized Sinkhorn algorithm from the POT library \citep{flamary2017pot} (leading to faster and more stable values) with a regularization parameter of 5.} (\textit{i.e.} from generated trajectories). As before, we measure both the distance to the training set and to the validation set and thus have two distinct metrics. The Wasserstein metric computation
assumes we can compute distances in the state space,
to this end we use the Euclidean distance with state coordinates
normalized to have the standard deviation equal to $1$. While this
is not directly applicable to vision-based observations, there are
plenty of techniques which can be used to compute vector embedding
for vision observations in RL/IL, e.g. \citet{atc, pi-sac}.

\textbf{Random Network Distillation (RND).} As proposed by \citet{burda2018exploration, wang2019random}, we define a support estimation metric by taking two randomly initialized networks, and train one to predict the \emph{random} output of the other one on the expert training set. The metric is then given by the MSE between the prediction network and the frozen one on trajectories generated by the agent. Details on the metric training are given in Appx.~\ref{app:rnd}.

\textbf{Imitation Return.} IRL algorithms recover a --learned-- reward function. We can use the corresponding learned return to select HPs. We compute this metric as it should model the goal optimized by the agent, although this goal is generally non-stationary. Moreover, the scale of the metric often depends on the HPs which could make it harder to compare the value of this metric between different training runs.

\textbf{Environment return.} This is the sum of rewards obtained during a full episode according to the environment reward function. We include it as the gold standard.

A desired property for all these metrics is to preserve the ranking of policies induced by the oracle environment return. A metric that would satisfy this property would yield an optimal HP selection. Note that this is a sufficient but not a necessary condition. We show in Appx~\ref{sec:policyranking}, using both
the Spearman rank correlation and the ROC-AUC of an additional task (classify good from poor policies), that the aforementioned metrics have good enough ranking properties to make them legitimate candidates for HP selection.

\subsection{Transfer scenario}
The HPs that work best in tasks for which a well-defined reward function is available can also be transferred to a new task.
We expect the success of this procedure to highly depend on the similarity between the source and target domains.

\begin{figure*}[!htb]
\centering
\begin{minipage}{0.75\linewidth}
  \centering
  \includegraphics[width=\linewidth]{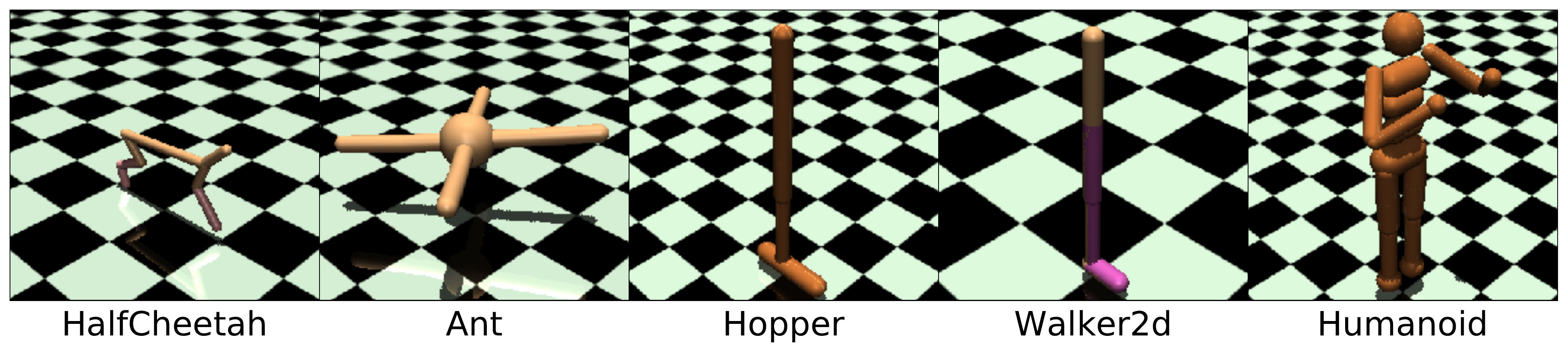}
  \includegraphics[width=\linewidth]{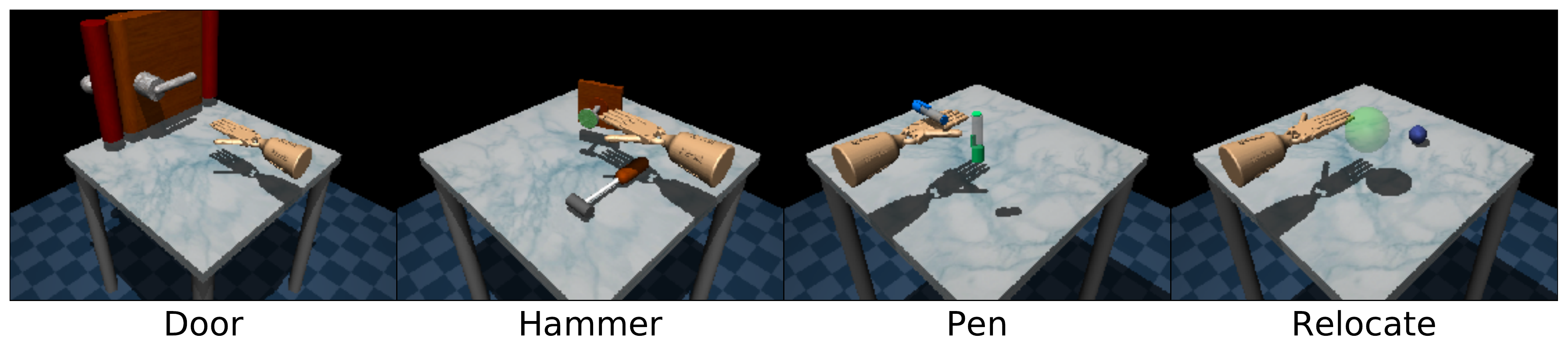}
\end{minipage}
\caption{Environments: OpenAI Gym (top) and Adroit (bottom).}
\label{fig:envs_adroit}
\end{figure*}

\subsection{Environments and data}
We focus on continuous-control benchmarks and consider five widely used environments from OpenAI Gym \citep{brockman2016openai}: \texttt{Hopper-v2}, \texttt{Walker2d-v2},
\texttt{HalfCheetah-v2}, \texttt{Ant-v2}, and \texttt{Humanoid-v2} and four manipulation tasks from Adroit \citep{Kumar2016thesis}: \texttt{pen-v0}, \texttt{relocate-v0}, \texttt{door-v0}, and \texttt{hammer-v0}. The Adroit tasks respectively consist in aligning a pen with a target orientation, moving an object to a target location, opening a door and hammering a nail. These two benchmarks bring orthogonal contributions. The former focuses on locomotion but has 5 environments with different state/action dimensionality. The latter, more varied in term of tasks, has an almost constant state-action space.
For the Gym tasks, we generate demonstrations with a Soft Actor-Critic (SAC) \citep{haarnoja2018soft} agent trained on the environment reward. For the Adroit environments, we use the ``expert'' datasets from the D4RL dataset \citep{fu2020d4rl}.

For the OpenAI Gym environments, we use 11 training trajectories and keep 5 additional held-out trajectories for validation. For the Adroit environments, 20 training trajectories are used as well as 5 validation trajectories. The number of training trajectories corresponds to what IL algorithms are typically designed for. We chose a low number of validation trajectories as demonstrations are generally expensive and one wishes to use as many as possible for training. 

\subsection{Algorithms}
We consider the three algorithms briefly described below. They use different approaches, from supervised learning, to Inverse RL via distribution matching, either through the primal or the dual form of a divergence. We consider them to be representative of different families of algorithms and thus good candidates to validate HP selection techniques. 

Detailed description of the algorithms can
be found in the original publications. We provide additional information
on our implementations in Appx.~\ref{app:setup}. We implemented our algorithms in the Acme framework \cite{acme}
using JAX \cite{jax} for automatic differentiation and Flax \cite{flax} for neural networks computation.

\textbf{Behavioral Cloning (BC, \citet{pomerleau1991efficient})} is the simplest approach to IL and relies on 
mapping the expert states to the expert actions in a supervised manner.
In contrast to the other algorithms we consider,
BC is an offline algorithm as it does not require any interactions
with the environment.

\textbf{Adversarial Imitation Learning (AIL, \citet{ho2016generative})} is a family of algorithms 
stemming from the seminal GAIL paper \citep{ho2016generative}.
The overall design uses a classifier to discriminate the expert state-action pairs from the agent ones, and an RL algorithm trains the policy to maximize
the confusion of this discriminator.
Our implementation is mostly similar to
Discriminator-Actor-Critic (DAC, \citet{kostrikov2018discriminator})
but uses different reward functions depending on the HPs.
See Appx.~\ref{app:ail} for details.

\textbf{Primal Wasserstein Imitation Learning (PWIL, \citet{dadashi2020primal})} uses an RL algorithm to minimize a greedy upper bound
on the Wasserstein distance between the state-action distributions of the expert and the agent.
For both  AIL and PWIL, we use Soft Actor-Critic (SAC, \citet{haarnoja2018soft})
to train the policy.

\subsection{Experimental design}\label{sec:setup-design}
For each algorithms and 5 different random seeds\footnote{The random seed fixes the episodes train/validation split.},
we sample 100 independent HP configurations from the HP sweeps
detailed in Appx.~\ref{app:setup}.
We then use them to train a total of 500 agents per algorithm-environment pair.  Online algorithms (AIL \& PWIL) are run for 1M environment steps while
BC is trained for 60k gradient steps. 
Each agent is evaluated 20 times throughout its training. At each evaluation, all the metrics are computed using 50 episodes\footnote{
We use only 10 episodes for the state divergence as it is more
computationally demanding.}.

We want to check whether the metrics
introduced in Sec.~\ref{sec:hp-selection}
can be used for the problem of HP selection.
To this end, we repeat the following experiment 20 times for each of the 5 seeds:
(1) we uniformly sample 25 HP configurations from the set of all 100 configurations,
(2) we take the final policies from the corresponding training runs,
(3) we select the best one according to the techniques
outlined in Sec.~\ref{sec:hp-selection}
and (4) we check how well it performs according to the
environment reward
This allows to simulate $5 \times 20$ practitioners that would run sweeps of size 25 and look for the best configuration possible.

We also investigate how the metrics described in Sec.~\ref{sec:hp-selection}
can be used for early stopping.
To this end, we repeat the above experiment but this time we select
the best policy not only from the fully trained ones, but also
include the partially trained policies
from the corresponding training runs.

Furthermore, we evaluate whether the performance on another task
with a well-defined reward function can be used to select HPs.
To this end, we choose a set of validation environments
and repeat the experiment mentioned above but
this time selecting the HP configuration with the best average normalized\footnote{Rewards are normalized per task so that $0$ corresponds to a random policy and $1$ is the average
return in the demonstration set.}
return across the validation tasks. We then check how well it performs on the test environment.
\section{Experimental Results}
In this section, we try to answer the following questions:

\textbf{1.} Can we effectively choose an IL
    algorithm and its HPs without access to the true reward function?\\
\textbf{2.} Which of the proxy metrics defined in Sec.~\ref{sec:hp-selection} is the best? \\
\textbf{3.} Is early stopping important in IL algorithms? \\
\textbf{4.} Do HPs transfer well between different environments?
\subsection{Using other metrics to select HPs}\label{subsec:metrics}

\begin{figure*}[!htb]
\centering
\includegraphics[width=.94\linewidth]{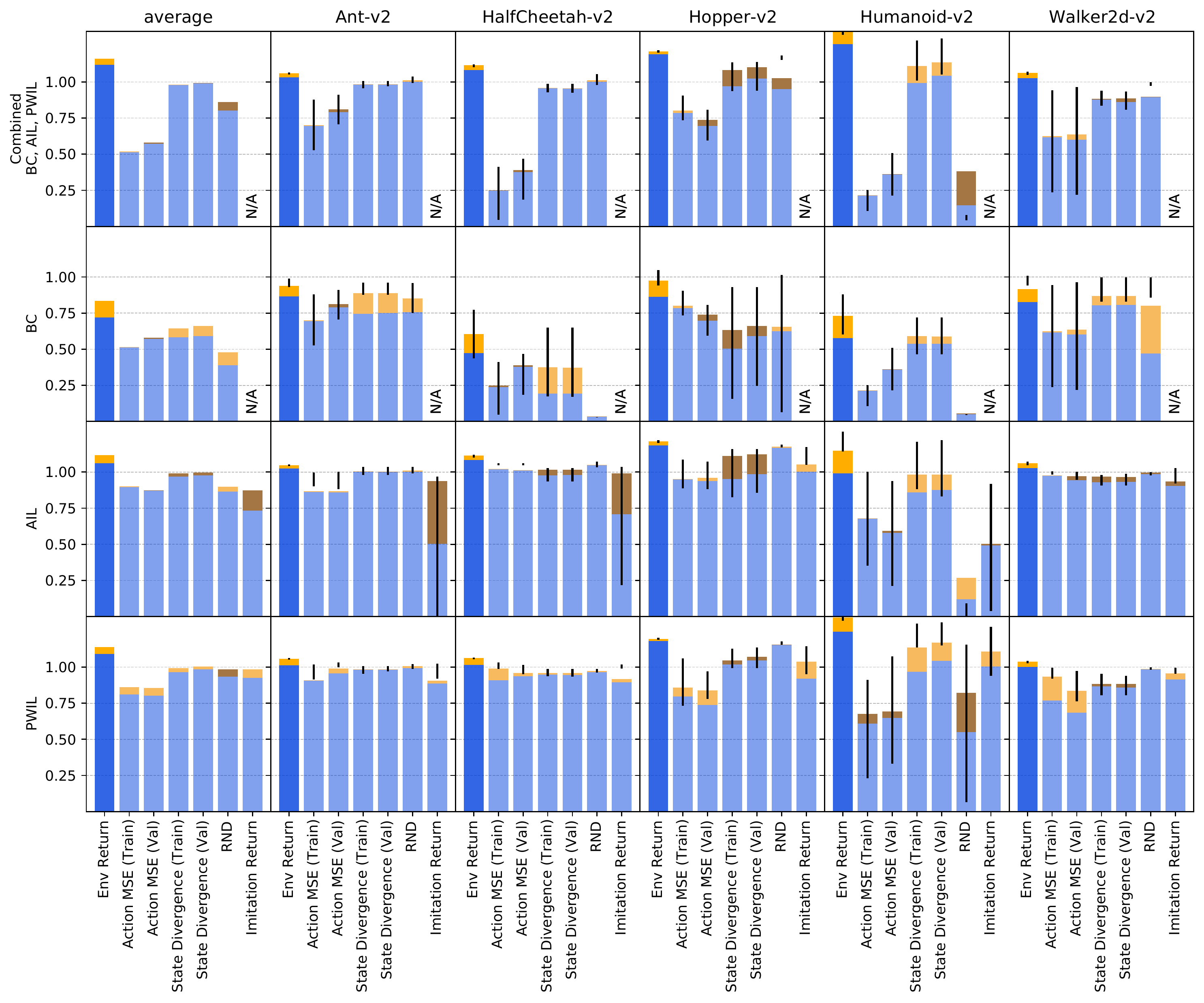}
\caption{
The episode return achieved by different algorithms if the
HP selection and early stopping is performed using a proxy metric for OpenAI Gym tasks.
Each subplot corresponds to a different algorithm and a different environment
with the first column showing the results averaged across environments.
The first row corresponds to the case when we choose HPs as well as the IL algorithm used
based on the given metric.
Episode returns are rescaled for each environment so that $0$ corresponds to a random policy and $1$
to the average episode return in the demonstration set. The lower (blue) part of each bar shows
the episode returns in the case of no early stopping and the full bar (blue and yellow)
shows the performance when using early stopping with the same metric as for the HP selection.
The vertical lines show the 25-th and 75-th percentile of the episode return
across reruning the whole HP selection process
as described in Sec.~\ref{sec:setup-design} for the early stopping case.
Brown color in the upper part of the bar means that the algorithm performs better without early stopping
and shows how much performance is lost by using it.
Each bar shows the mean performance across running the HP selection process 100 times.}
\label{fig:scores-mujoco}
\end{figure*}

\begin{figure*}[!htb]
\centering
\includegraphics[width=0.80\linewidth]{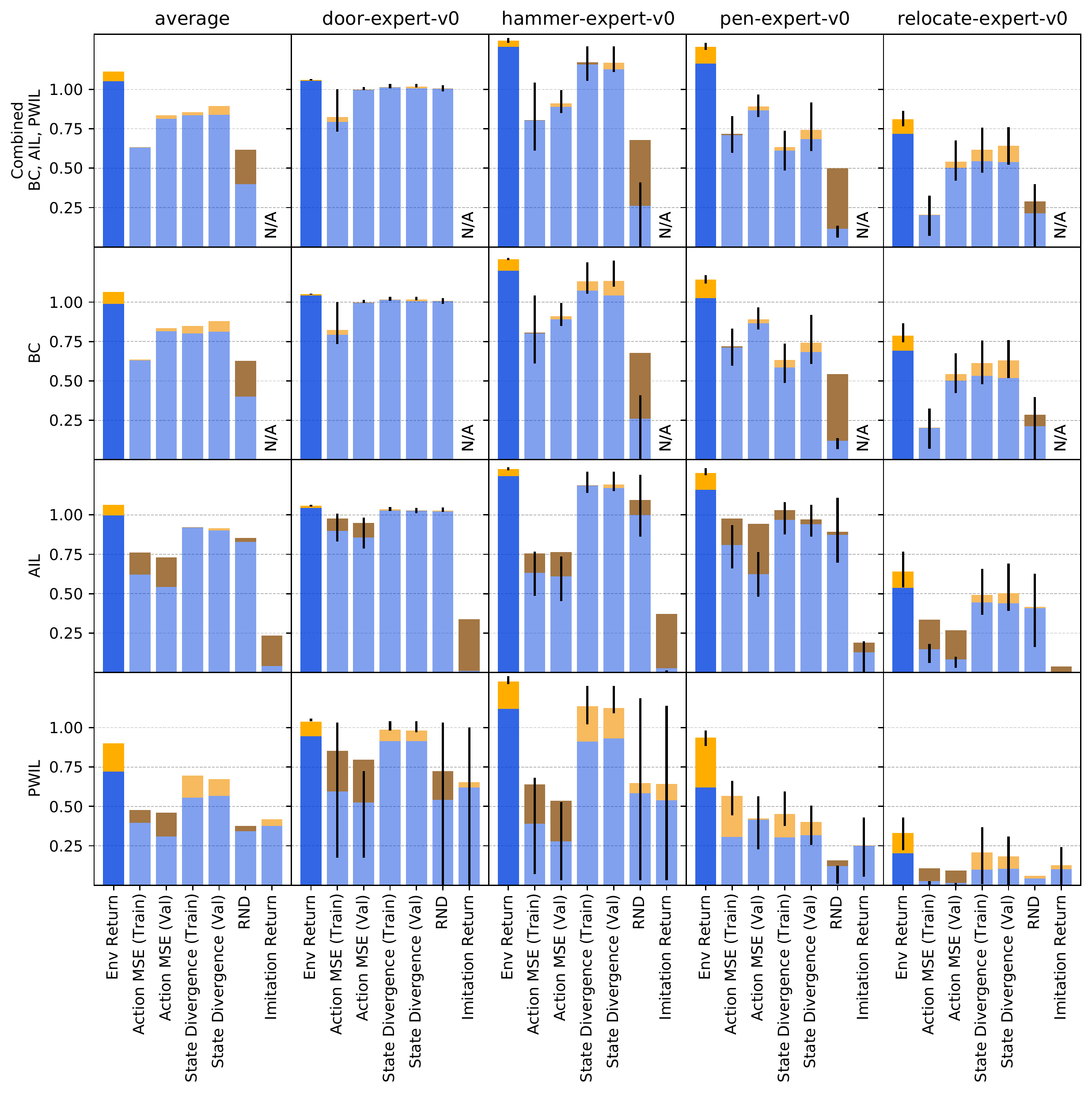}
\caption{
The episode return achieved by different algorithms if the
HP selection and early stopping is performed using a proxy metric for Adroit environments.
See the caption for Fig.~\ref{fig:scores-mujoco} for the detailed description of this plot.}
\label{fig:scores-adroit}
\end{figure*}

Fig.~\ref{fig:scores-mujoco} and Fig.~\ref{fig:scores-adroit} show
how the performance of different algorithms varies as we change the metric used for HP selection
on OpenAI Gym and Adroit environments respectively.

By looking at the first bar in each subplot (performance for HPs selected on the episode return),
we can see that all algorithms, including BC\footnote{
Many recent publications in IL (e.g. \citet{ho2016generative})
\emph{subsample} demonstrations
by including only every $n$-th state-action pair to
make the task harder for BC.
We did not follow this practice
as it has little justification from the practical point of view.
},
can achieve returns comparable to or even better than those of
the expert on almost all environments if the HP selection and early stopping are performed
using the oracle episode return.
In some cases, the policies obtained this way can even perform
significantly better than the expert (e.g. BC on \texttt{pen-expert},
Fig.~\ref{fig:scores-adroit}). This should not happen
as all the algorithms are trying to mimic the expert behaviour and we argue that this is an artifact of the policy selection process.

\textbf{How much do we lose by choosing an IL algorithm and HPs using a proxy metric?}
The top-left subplot in Fig.~\ref{fig:scores-mujoco} and Fig.~\ref{fig:scores-adroit} shows the performance averaged
across the tasks in a given suite when
the HPs \emph{as well as} the IL algorithm are
chosen using the proxy metrics.
The selection by state divergence achieves episode returns
similar to that of the demonstrator on both environment
suites.
This is a \textit{very positive result}, suggesting that it is,
in practice, possible to select HPs without access to a reward function and still obtain a policy that performs well at the task.

On the other hand, the big gap between the performance of all three algorithms on Adroit~(first
column in Fig.~\ref{fig:scores-adroit}) with HPs
selected on the return \textit{vs.} proxy metrics shows that
the research practice of selecting HPs based on the return can lead to significant overestimation
of an algorithm's performance when no reward function is available. This may limit the applicability of some IL algorithms to practical problems.

\textbf{Which proxy metric is the best?}
The state divergence performs best for all algorithms.
Action MSE performs worse than the state divergence but still
achieves at least 75\% of the expert score
on some algorithm-environment pairs.
The inferior performance of action MSE
is expected as it is a fully offline metric
unaware of the system dynamics.
Whether the metrics are computed against the set of training demonstrations
or a held-out set of demonstrations makes little difference
in our experiments
and suggests that it might be better to use all available demonstrations for training\footnote{
For some algorithm-environment pairs,
the train versions of some metrics performs better than the validation ones.
We suspect that it is due to
the training sets being bigger (11 trajectories for OpenAI Gym and 20 for Adroit) than the validation sets (5 trajectories).
Moreover, except for BC and action MSE,
the metric used to select HPs is not the metric being optimized,
so it may be less crucial to have a validation set.}.


The metric given by the average RND score on the episode performs well to select HPs
on OpenAI Gym environments but is outperformed by the state divergence.
Perhaps suprisingly, the metric is not as performant on Adroit. This suggests that this support-estimation metric might have to be adapted for environments with more stochasticity in initial states.
Imitation return (the sum of learned-reward collected in an episode), as defined by in AIL and PWIL also perform poorly and
is usually worse than action MSE with the exception
of PWIL on OpenAI Gym where both approaches perform similarly.
The inferior performance of the imitation return for AIL
can be explained by the fact that the reward function introduced in the
algorithm is non-stationary. Therefore comparing the values
from different training runs or different timesteps might be misleading. 
Concerning PWIL, we suspect that its inferior performance on Adroit compared
to the state divergence metrics is caused by the fact that
the upper-bound on the Wasserstein distance introduced by the algorithm
is not tight.


\textbf{Is early stopping important?}
The upper yellow (resp. brown) parts of the bars in Fig.~\ref{fig:scores-mujoco} and Fig.~\ref{fig:scores-adroit}
show how much is gained (resp. lost) by using early stopping based on the
same metric as for the HP selection. We can see that early stopping
almost always improves performances if a reliable metric
(\textit{i.e.}, state divergence) is used and the task was not already
almost completely solved without it.
The magnitude of the gain
depends heavily on the environment and the algorithm, in particular
we found early stopping to be particularly helpful for PWIL on Adroit environments.

\textbf{Can we use metrics to choose the algorithm?}
The first row in Fig.~\ref{fig:scores-mujoco} and Fig.~\ref{fig:scores-adroit}
show the performance when we use the metrics to choose not only HPs
but also the IL algorithm.
We observe that when using the state divergence as the metric to select the algorithm,
we achieve comparable performance to
that of the best algorithm.
Furthermore, selecting the algorithm based on action MSE
results in similar performance as BC even if it is not the best
algorithm. This may be due to the fact that
BC optimizes the very same criterion, hence
it is more likely to have the best results according to that metric.
We provide additional evidence for this hypothesis in Appx.~\ref{app:appendix-algo-select}.


 
\subsection{Hyperparameter selection by transfer}
\begin{figure*}[!htb]
\centering
\begin{minipage}{\linewidth}
    \centering
  \includegraphics[width=0.94\linewidth]{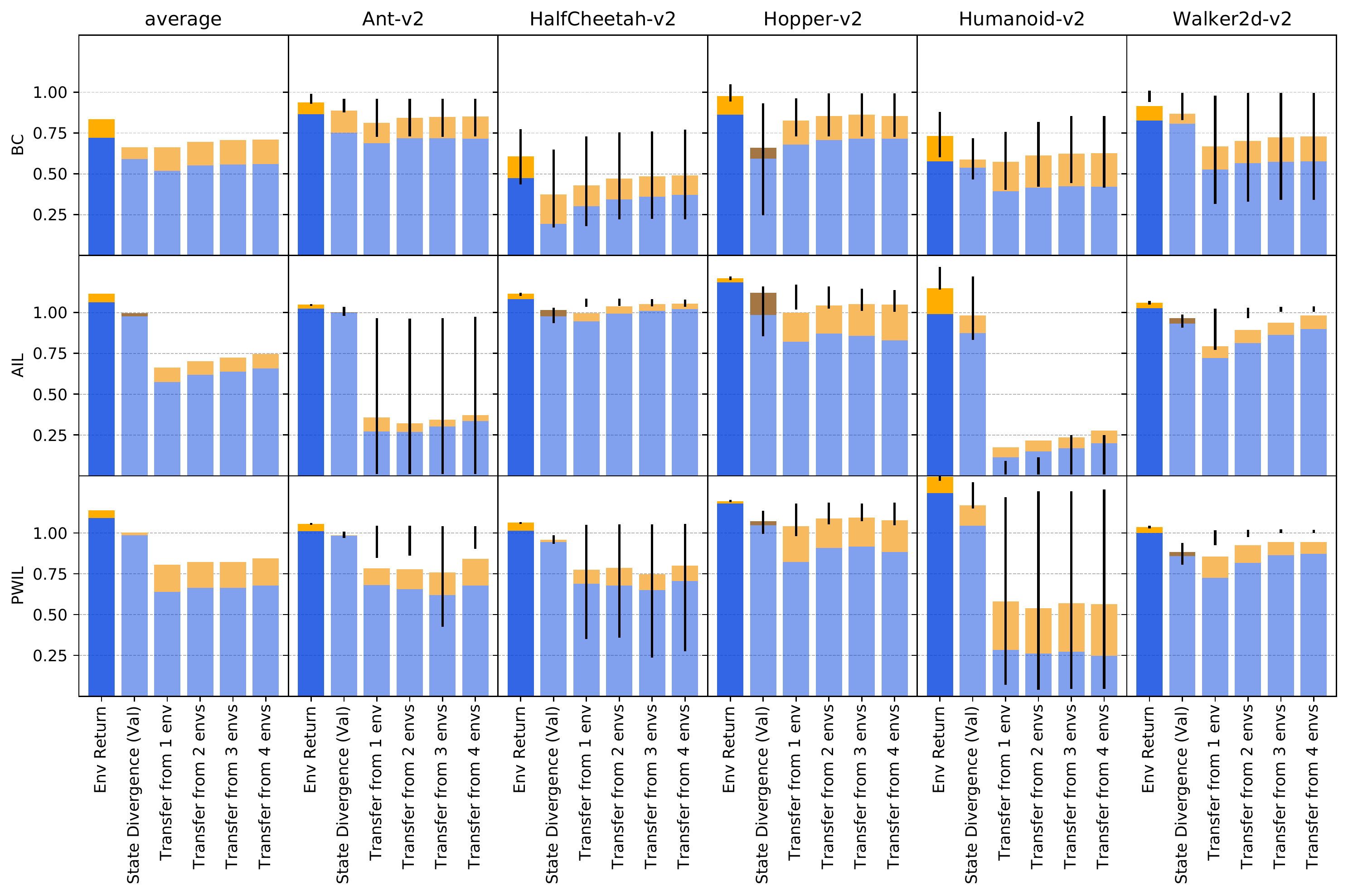}
\end{minipage}
\caption{The episode return achieved by different algorithms if the
HP selection is performed using transfer for OpenAI Gym environments.
HPs are selected by choosing the HP configuration
which performs best on a set of other tasks from the same benchmark
(See Sec.~\ref{sec:setup-design} for the details).
The results are averaged across all possible choices of the
validation environments.
Different bars in each subplot correspond to
using a different number of environments to select HPs.
We also include the selection based on the episode return and the
state divergence for comparison.
Early stopping is performed using the state divergence (validation)
regardless of the HP selection metric.
See the caption for Fig.~\ref{fig:scores-mujoco}
for additional information on this figure.
}
\label{fig:transfer-mujoco}
\end{figure*}

\begin{figure*}[!htb]
\centering
\begin{minipage}{0.80\linewidth}
    \centering
  \includegraphics[width=\linewidth]{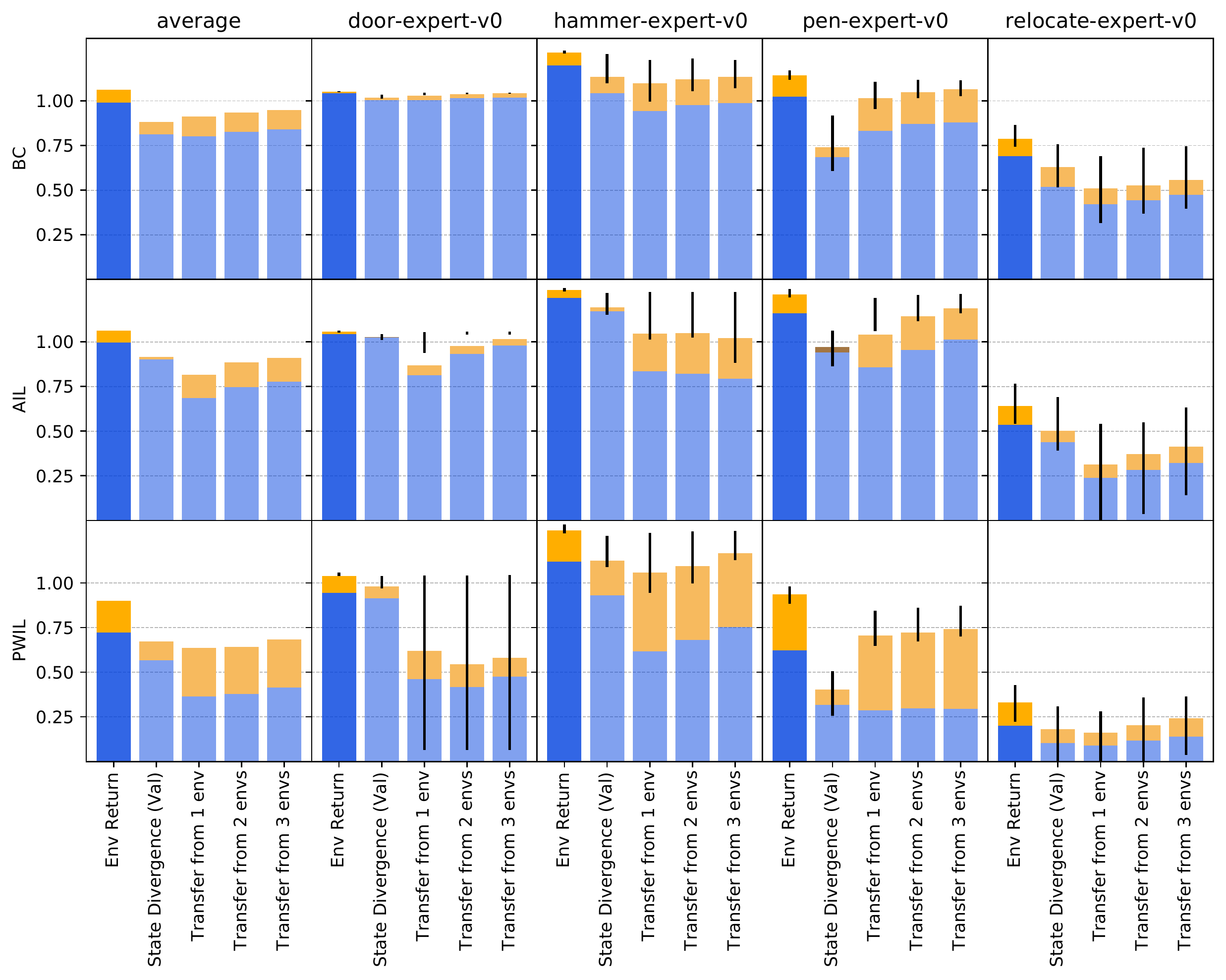}
\end{minipage}
\caption{The episode return achieved by different algorithms if the
HP selection is performed using transfer for Adroit environments.
See Fig.~\ref{fig:transfer-mujoco} for additional information
on this plot.}
\label{fig:transfer-adroit}
\end{figure*}

\begin{figure*}[!htb]
\centering
\begin{minipage}{\linewidth}
  \centering
  \includegraphics[width=0.43\linewidth]{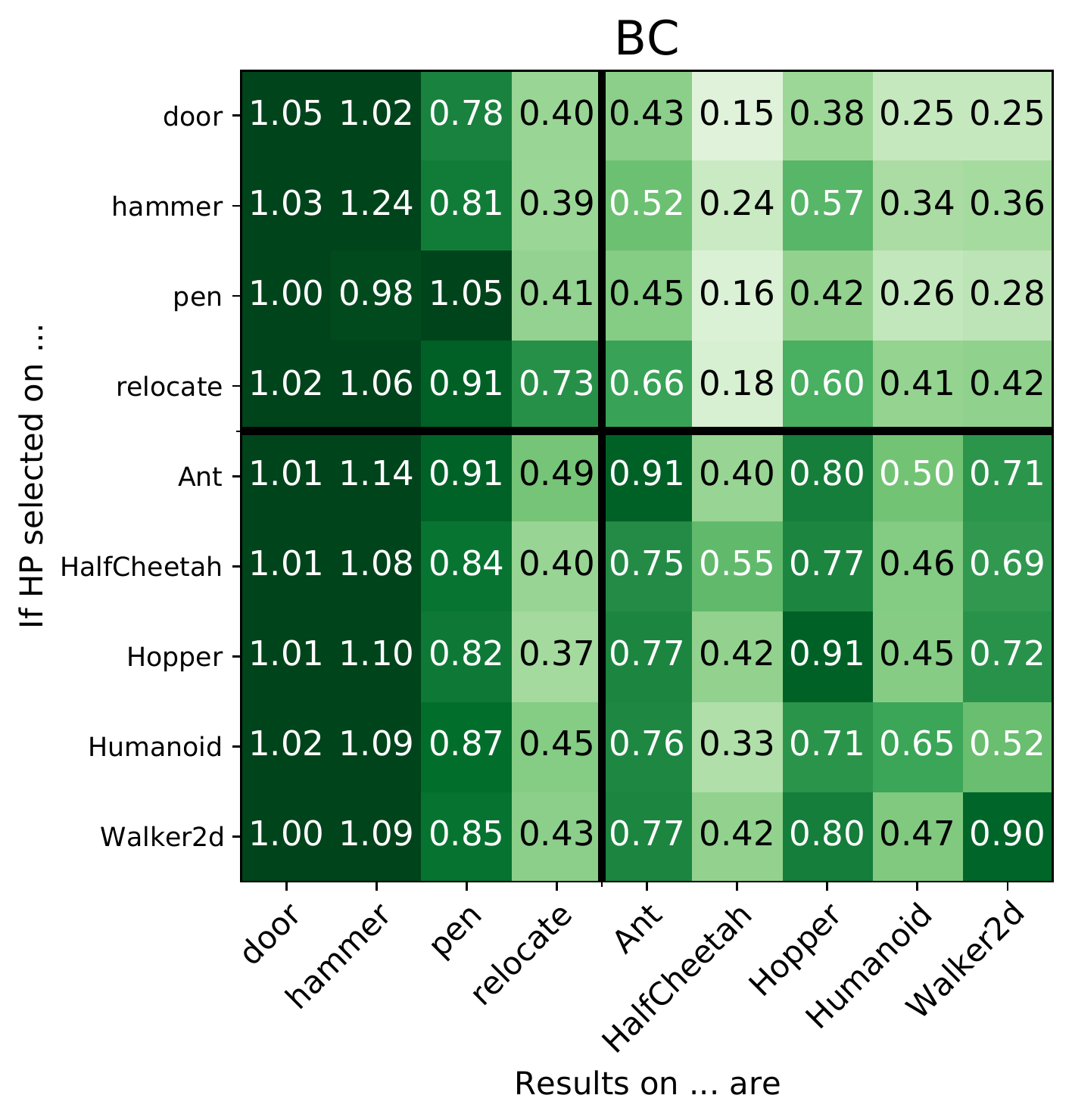}
  \includegraphics[width=0.43\linewidth]{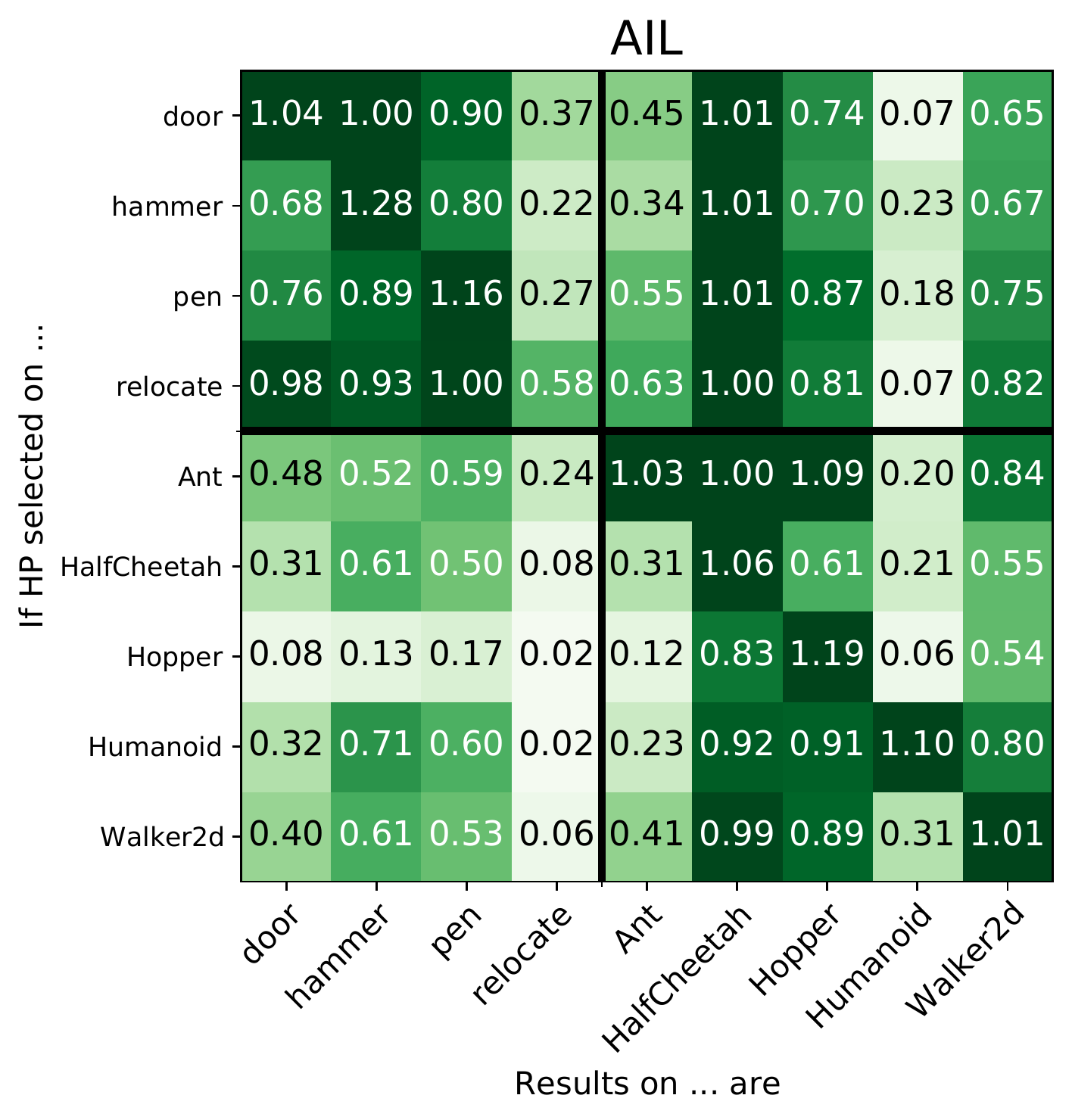}
  \includegraphics[width=0.43\linewidth]{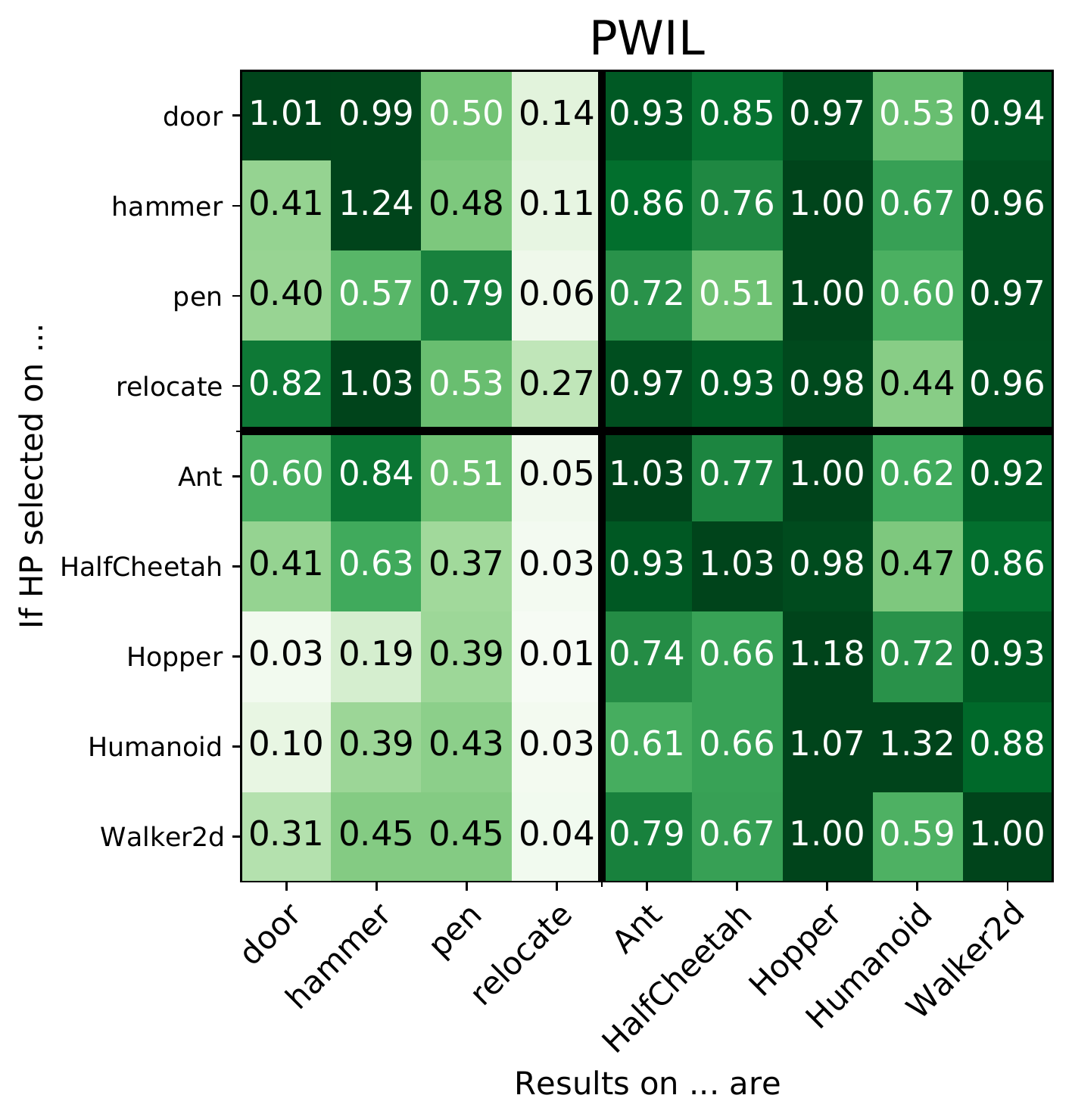}

\end{minipage}%

\caption{HP transfer results for individual validation-test environments pairs.
Rows correspond to different validation environments and columns to different
test environments.
Early stopping is performed using the state divergence on the validation
set of demonstrations.}
\label{fig:transfer-wasserstein}
\end{figure*}

As an alternative to the metrics investigated in the previous section,
we could select HPs which work best on a similar task(s)
that actually has a well-defined reward function.
Fig.~\ref{fig:transfer-mujoco} and~\ref{fig:transfer-adroit} show the performance of different algorithms when we choose HPs using a set of validation environments and then
test them on another environment from the same benchmark (\textit{i.e.},
we do not consider the transfer from OpenAI Gym to Adroit here).
While we could use transfer for early stopping
(\textit{i.e.}, select the timestep which worked best for the validation environments),
we have noticed that it almost always resulted in worse performance than
no early stopping and therefore we chose to
use state divergence for
early stopping in the transfer experiments.

Transfer performance consistently improves with the number of
validation environments used but
selecting HPs based on the state divergence on the task of interest
outperforms transfer
even if we validate HPs on all the other tasks in the given task suite.
The relatively poor transfer performance
can be caused by two factors: (1) different HP configurations
performing well on each task, or (2)
the stochasticity of the algorithm (\textit{i.e.}, the algorithm
producing different results when run twice with the same HPs).
Despite that, transferring HPs may still be preferred in some
situations as it does not require tuning HPs for each new task.
 
Fig.~\ref{fig:transfer-wasserstein} shows the transfer performance
for individual pairs of validation-test environments.
We observe that HPs transfer better within a suite,
but the HP transfer between OpenAI Gym and Adroit  often succeeds too. Especially, BC's HPs transfer very well from
Adroit to Gym but not the other way around.
Results also suggest that HPs transfer better to easier tasks
than to harder ones:
the two tasks on which the HP transfer performs worst
are \texttt{Humanoid} and \texttt{relocate} which are the hardest tasks
in their respective suites.

The results suggest also that BC enjoys better HP transferability
than AIL and PWIL which can be expected as it is
a much simpler algorithm based on supervised learning.

A general conclusion from our experiments is that IL algorithms are quite sensitive to their
HPs and may require per-environment HP tuning for optimal performance.
It is therefore important to compare IL algorithms not only in terms of their performance under optimal HPs but also in terms of their HP sensitivity and transferability.

%
\section{Related Work}
Hyperparameters selection is central to the performance of machine learning algorithms. In the context of supervised learning, it is common practice to select HPs on dedicated held-out data (validation) and subsequently estimate the performance of an algorithm on another set of held-out data (test). Previous work has looked into improving the selection of the best configurations using grid search \citep{lecun2012efficient}, random search \citep{larochelle2007empirical,bergstra2012random} or population based strategies \cite{jaderberg2017population}.

In the context of reinforcement learning, the notions of training and validation can be conflated. This is the case if the goal is to design a policy that performs well in a single environment \citep{silver2016mastering,tesauro1995temporal,vinyals2017starcraft}. However, RL agents are typically evaluated on their ability to learn on multiple environments \citep{bellemare2013arcade,brockman2016openai,cobbe2020leveraging} with a single set of HPs. \citet{henderson2018deep} highlights that poor evaluation protocols make RL algorithms hard to reproduce. For instance, although code level optimizations often explain a lot of the RL algorithms performance, they tend to be omitted in most of the recent RL publications~\cite{engstrom2020implementation,andrychowicz2020matters}, making it hard for practionners to compare and reproduce.

Imitation Learning adds another level of algorithmic complexity to the RL setting since the reward function is not available. Therefore, the learning of a reward function, which introduces its own set of extra HPs, is intertwined with the learning process of a direct RL agent \cite{finn2016guided,ho2016generative,kostrikov2018discriminator}. Although previous work has identified measures of similarity not based on the reward functions \cite{dadashi2020primal,ghasemipour2020divergence,ke2019imitation}, this work is, as far as we know, the first to propose a principled evaluation protocol to select HPs not based on the true but supposedly unknown reward function.

\citet{paine2020hyperparameter} recently highlighted the same problem occurring in offline RL \cite{lagoudakis2003least,ernst2005tree,riedmiller2005neural,lange2012batch,levine2020offline}, where HPs are usually selected on the performance of the agent on the online environment (although the core constraint of offline RL prevents interactions with the environment). Some recent benchmarks \citep{gulcehre2020rl,fu2020d4rl} also propose evaluation protocols
where HPs are selected by tuning on only a subset of environments.

\section{Summary/Conclusion}
In this work, we highlighted a major flaw in current evaluation protocols of IL methods. Although the promise is to design agents learning from demonstrations, the standard practice is to select agents on the reward of the task. In order to align research progress with the problem it attempts to solve, we advocate for a new evaluation protocol, where the HP selection is based on criteria available in the IL setting.
We investigated multiple proxies to the environment return for HP selection and early stopping. We evaluated, on 9 continuous control tasks, model selection using proxy metrics or through transfer.
We demonstrated the brittleness of classical algorithms when the HP selection cannot be performed on the unknown environment return. We also showed that it is possible to select good HPs by estimating
the divergence between the distribution of states encountered by the demonstrator
and the agent. 
This work opens the interesting question of new proxy metrics design, that can adapt to harder IL settings including suboptimal demonstrations, partial observability or visual-based inputs.



\section*{Acknowledgments}
We thank Lucas Beyer, Johan Ferret and Nino Vieillard for their feedback on earlier versions of the manuscript.
\clearpage
\bibliography{biblio}
\bibliographystyle{icml2021}

\clearpage
\appendix
\section{Experimental details}\label{app:setup}
All experiments use MLP networks (optionally with dropout \citep{dropout})
and Adam \citep{adam} optimizer.
Some experiments use observation normalization which means that the
observation coordinates are rescaled to have the mean equal $0$
and standard deviation equal $1$.
The normalization statistics are computed using the dataset
of expert trajectories.

Both, AIL and PWIL use SAC \citep{haarnoja2018soft}
with an entropy constraint \citep{haarnoja2018soft2}
to train the policy
and optionally use absorbing states
introduced in DAC \citep{kostrikov2018discriminator}.
When evaluating SAC policies,
we decrease the stochasticity of the policy by sampling each actions 5 times
and averaging the sampled actions.

\subsection{Behavioral Cloning}

The HP sweep for BC can be found in Table~\ref{table:sweep-bc}.

\begin{table}[h]
  \caption{Hyperparameter sweep for BC.}
  \label{table:sweep-bc}
  \centering
  \begin{tabular}{clr}
    \toprule
    Hyperparameter & Possible values \\
    \midrule
    gradient updates & 6000 \\
    batch size & 256 \\
    learning rate & $10^{-5}$, $10^{-4}$ \\
    weight decay coefficient & 0, 0.01, 0.1 \\
    observation normalization & True, False \\
    policy output & deterministic \\
    \# of policy hidden layers & 1, 2, 3\\
    hidden layer size & 16-256 \\
    activation function & ReLU, tanh \\
    input dropout rate & 0, 0.15, 0.3 \\
    hidden dropout rate & 0, 0.25, 0.5 \\
    \bottomrule
  \end{tabular}
\end{table}

\subsection{Adversarial Imitation Learning}\label{app:ail}

The HP sweep for AIL can be found in Table~\ref{table:sweep-ail}.
The AIL experiment samples the policy reward function from
the following options \citep{kostrikov2018discriminator, ghasemipour2020divergence}:
\begin{itemize}
\item  $\log D(s,a)$,
\item $-\log (1-D(s,a))$,
\item $\log D(s,a)-\log (1-D(s,a))$.
\end{itemize}
To prevent the discriminator saturation,
we subtract the entropy of the discriminator output (treated as a Bernoulli distribution)
from the discriminator loss.

\begin{table}[h]
  \caption{Hyperparameter sweep for AIL.}
  \label{table:sweep-ail}
  \centering
  \begin{tabular}{clr}
    \toprule
    Hyperparameter & Possible values \\
    \midrule
    discriminator training & \\
    \# of gradient steps per env step & 1 \\
    training batch size & 256 \\
    \# of hidden layers & 2 \\
    hidden layer size & 256 \\
    activation function & ReLU \\
    learning rate & $10^{-6}$--$3\cdot10^{-4}$ \\
    entropy coefficient & $10^{-4}$--$10^{-2}$ \\
    \midrule
    RL reward function & See~App.~\ref{app:ail}  \\
    observation normalization & True, False \\
    absorbing states & True, False \\
    RL training & See Table~\ref{table:sweep-sac}. \\
    \bottomrule
  \end{tabular}
\end{table}

\begin{table}[h]
  \caption{Hyperparameter sweep for SAC.}
  \label{table:sweep-sac}
  \centering
  \begin{tabular}{clr}
    \toprule
    Hyperparameter & Possible values \\
    \midrule
    environment steps & 1M \\
    replay buffer size & 1M \\
    gradient steps per env step & 1 \\
    learning rate & $3\cdot10^{-5}$ -- $1\cdot10^{-3}$ \\
    batch size & 128, 256, 512 \\
    discount factor & 0.9, 0.97, 0.99 \\
    target network coefficient & 0.001 -- 0.03 \\
    reward scale & 0.01 -- 1 \\
    target entropy & -0.5 per dimension \\
    \midrule
    policy & \\
    action distribution & Gaussian + tanh \\
    \# of hidden layers & 2 \\
    hidden layer size & 256 \\
    activation function & ReLU \\
    \bottomrule
  \end{tabular}
\end{table}

\subsection{Primal  Wasserstein  Imitation  Learning}

The hyperparameter sweep for PWIL can bew found in Table~\ref{table:sweep-pwil}.
PWIL prefills the SAC replay buffer with a number of transitions
from the expert dataset\footnote{The reward for these transitions is set to $\alpha$ where
$\alpha$ is the coefficient used in the PWIL reward definition \citep{dadashi2020primal}.
If the number of requested transitions is higher than the size of the expert dataset,
we include multiple copies of each transition.}.

\begin{table}[h]
  \caption{Hyperparameter sweep for PWIL.}
  \label{table:sweep-pwil}
  \centering
  \begin{tabular}{clr}
    \toprule
    Hyperparameter & Possible values \\
    \midrule
    observation normalization & True, False \\
    absorbing states & True, False \\
    reward coefficient $\alpha$ & 0.5, 1, 5, 10 \\
    reward coefficient $\beta$ & 0.5, 1, 5, 10 \\
    \# of expert transitions in the buffer & 0, 5k, 50k \\
    RL training & See Table~\ref{table:sweep-sac}. \\
    \bottomrule
  \end{tabular}
\end{table}

\subsection{Random Network Distillation metric}\label{app:rnd}
This metric uses a first frozen feed-forward network $f$ with layer sizes $(128, 128, 128, 128)$ as a target and a second learnable network $\hat{f}$ with layer sizes $(128, 128)$ to predict the output of the former.
The latter is trained with Adam and a learning rate of $0.001$ to minimize the mean-squared error for 100 epochs on the demonstration dataset.
The corresponding metric is given by $\exp(-||f(obs)-\hat{f}(obs)||)$.

\section{Additional results}

\subsection{Ranking property of the proposed metrics}
\label{sec:policyranking}

Whether a metric is good for HP selection mostly depends on its capacity to preserve the ranking
between policies induced by the oracle return. 
Assuming a set of policies whose expected return is evenly distributed over an interval, we can introduce two tasks that can serve as an intrinsic evaluation of the proxy metrics.
The first task aims at ranking all the policies according to the chosen metric. The performance of this ranking is given by the Spearman correlation with the ranking provided by the oracle return.
The second task aims at differentiating the set of good policies, i.e. policies whose oracle return is above a threshold (chosen to be 75\% of the expert performance),  
from the set of poor policies: we compute the probability that the chosen metric is lower for a policy sampled randomly from the set of good policies than for a policy sampled randomly from the set of poor policies. This probability also corresponds to the ROC-AUC of a classifier that would be based on the same metric.

To create a set of policies, we train some agents using one of the IL algorithms configured as described in Sec.~\ref{app:setup} and collect for each HP configuration the policies from different stages of training. The resulting set of policies is highly imbalanced in terms of environment returns.  To get policies evenly distributed in terms of environment returns, we finally rebalance the set of policies.
The two tasks are repeated for each set of policies obtained using BC, AIL and PWIL and the corresponding scores are averaged.

\begin{figure*}[!htb]
\centering
\begin{minipage}{\linewidth}
  \centering
  \includegraphics[width=\linewidth]{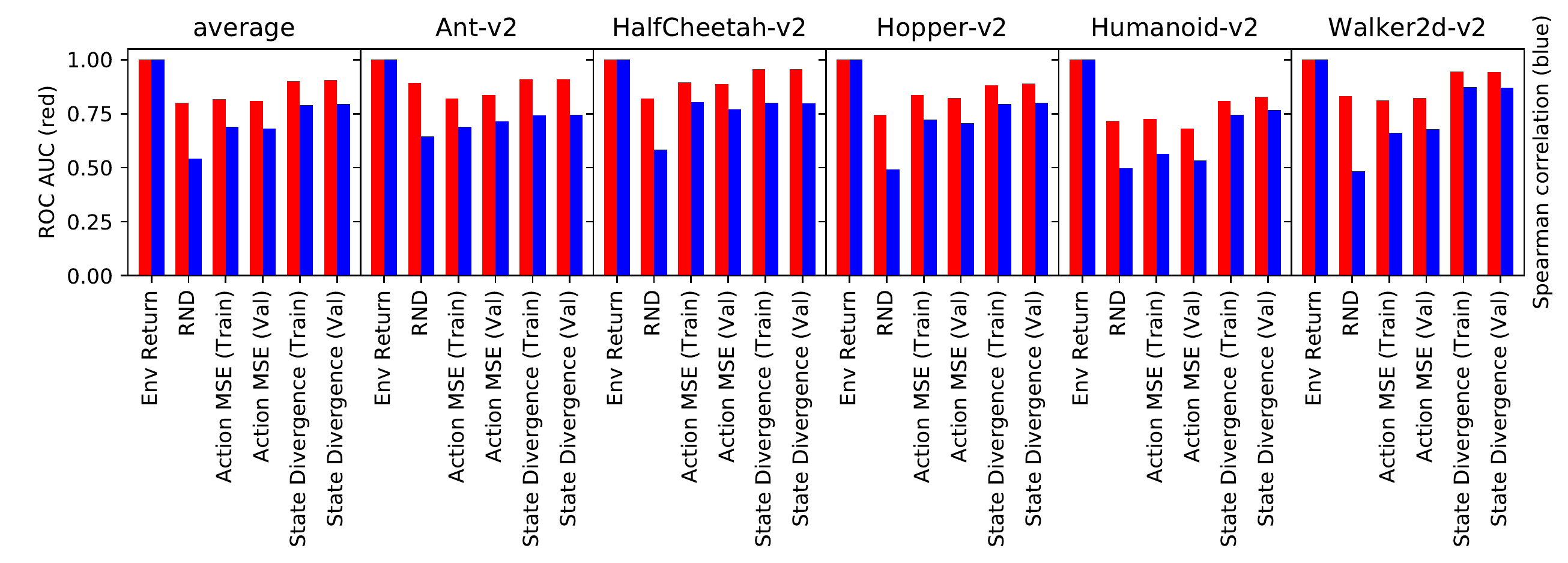}
  \includegraphics[width=\linewidth]{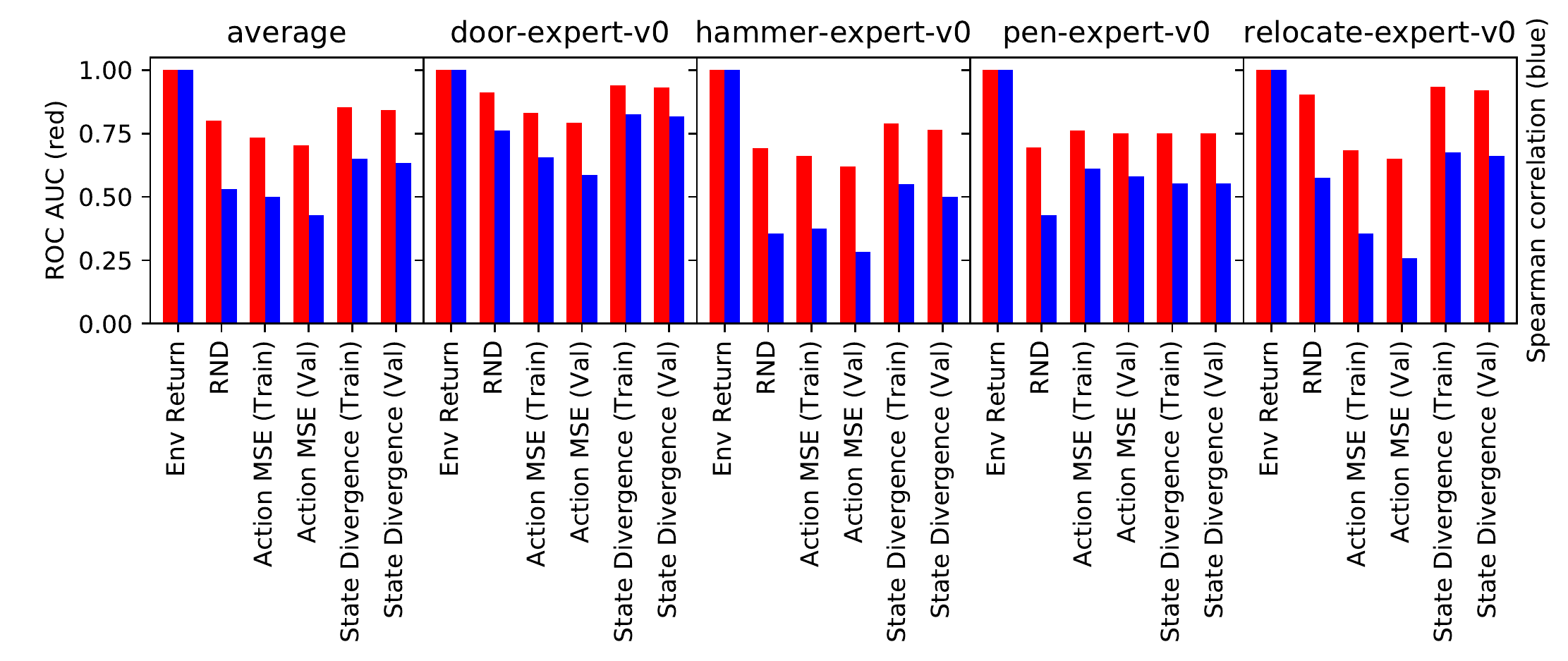}
\end{minipage}%
\caption{ROC-AUC (red) and Spearman correlation (blue) of the proposed metrics with respect to the oracle
return on a set of Mujoco and D4RL adroit environments.}
\label{fig:rocaucspearman}
\end{figure*}

Fig.~\ref{fig:rocaucspearman} clearly shows the metrics somewhat preserve the ranking induced by the oracle return. The best metric on those two ranking tasks is the state divergence, suggesting this could be the most suitable metric for HP selection.





\subsection{Selecting the algorithm}
\label{app:appendix-algo-select}

\begin{figure*}[!htb]
\centering
\includegraphics[width=\linewidth]{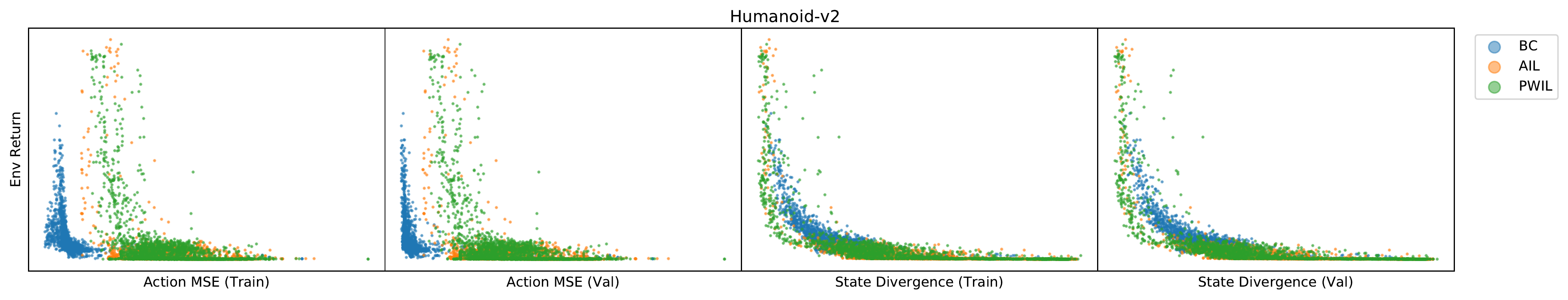}
\caption{Values of the environment return and of the action MSE and state divergence metrics for all the models (across HPs and training steps) of the three different imitation learning algorithms.}
\label{fig:scatter-humanoid}
\end{figure*}

The first row of Fig.~\ref{fig:scores-mujoco} gives the episode returns for the different proposed metrics when the imitation learning algorithm is itself considered as an hyperparameter. The results suggest the action MSE metric favors BC even when BC is actually not the best algorithm.
We provide in Fig.~\ref{fig:scatter-humanoid} an evidence of this hypothesis through a detailed view of all the partially-trained models from different HP configurations
for each of the three imitation learning algorithms. The models with best action MSE correspond to models learned with BC although they are clearly not the best models in this case.

\subsection{Transfer}
\begin{figure*}[!htb]
\centering
\begin{minipage}{\linewidth}
  \centering
  \includegraphics[width=0.33\linewidth]{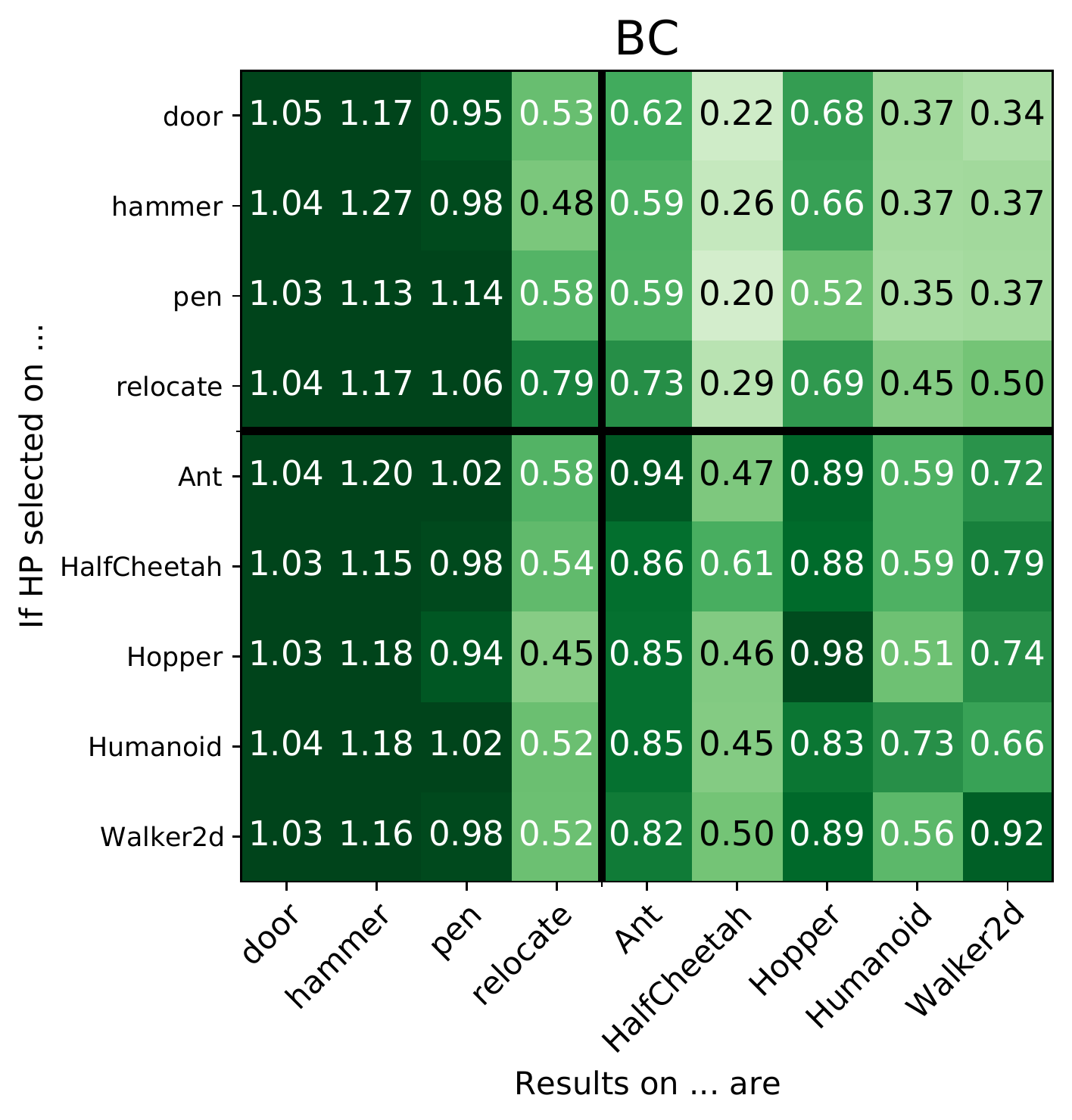}
  \includegraphics[width=0.33\linewidth]{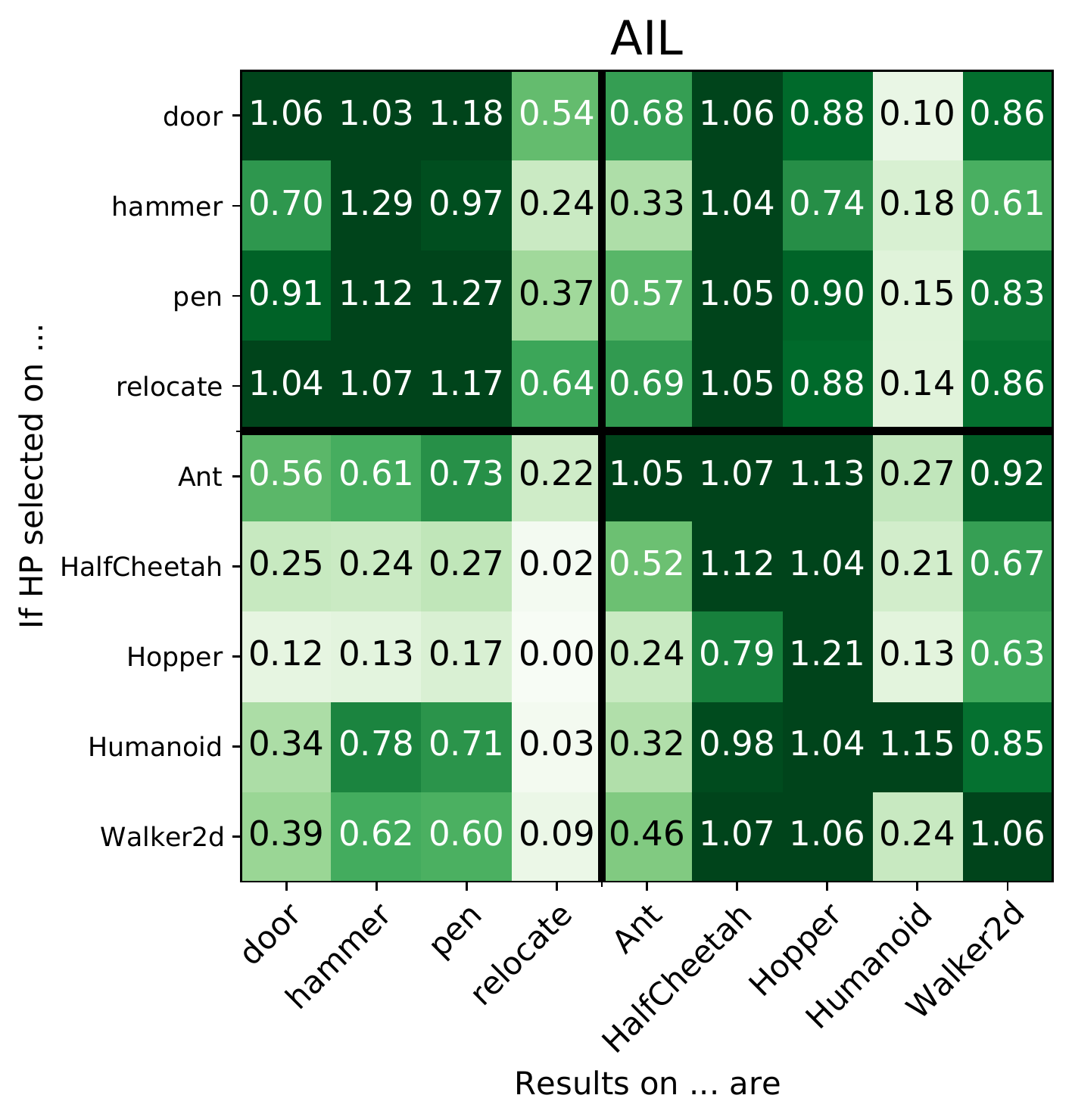}
  \includegraphics[width=0.33\linewidth]{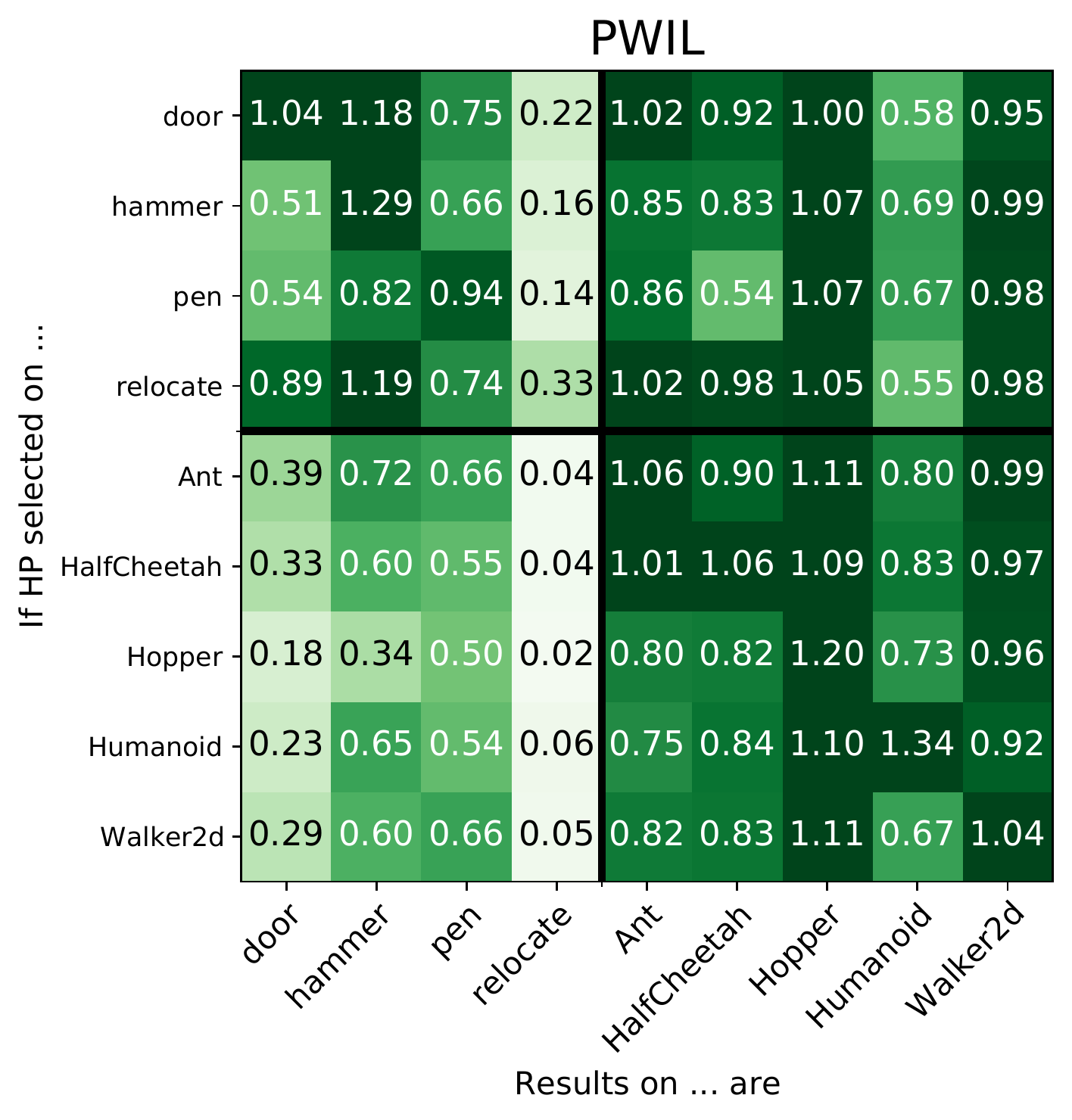}

\end{minipage}

\caption{HP transfer results for individual validation-test environments pairs.
Rows correspond to different validation environments and columns to different
test environments.
Early stopping is performed using the oracle return.}
\label{fig:transfer_detail_oracle}
\end{figure*}

\begin{figure*}[!htb]
\centering
\begin{minipage}{\linewidth}
  \centering
  \includegraphics[width=0.33\linewidth]{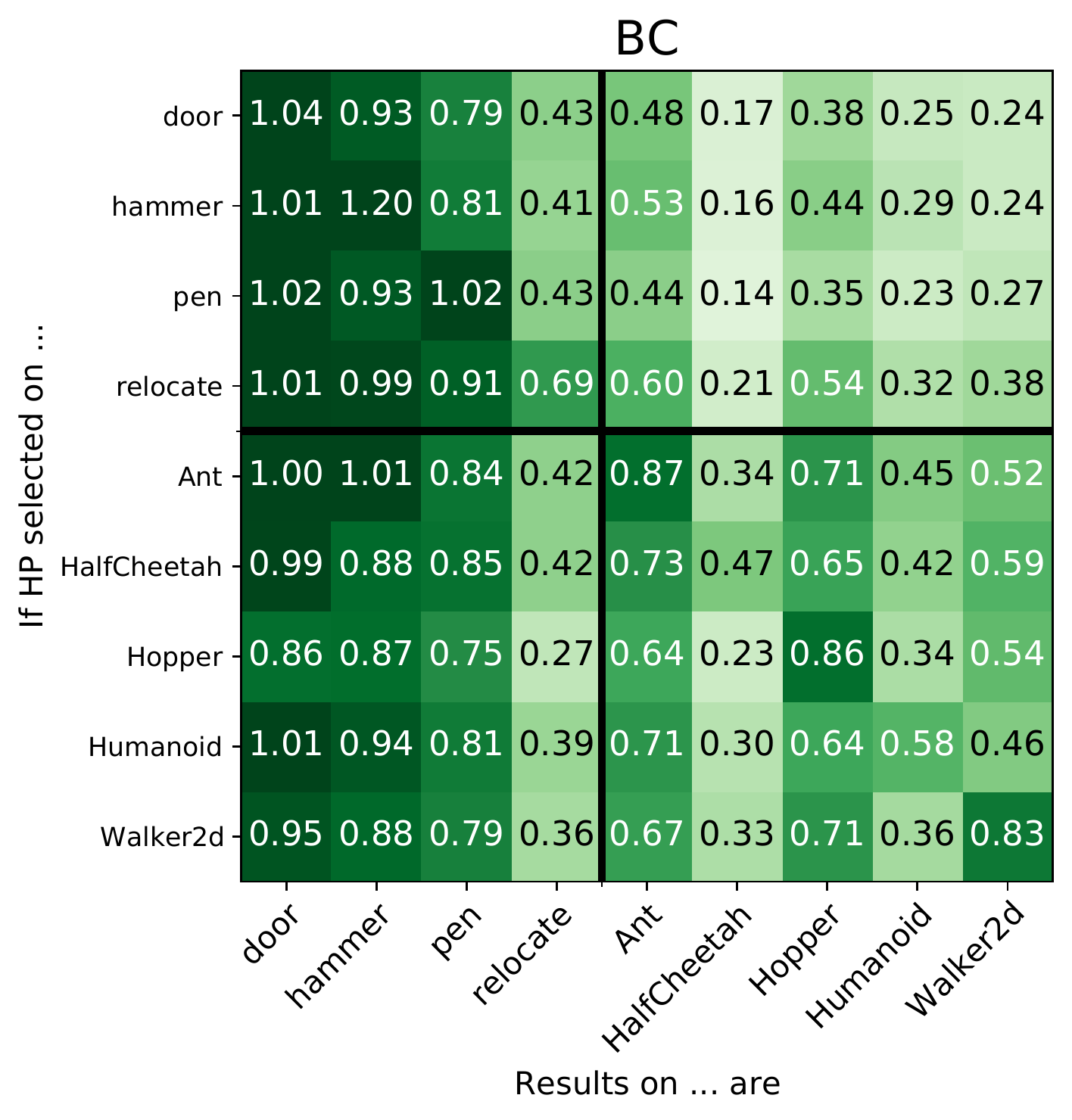}
  \includegraphics[width=0.33\linewidth]{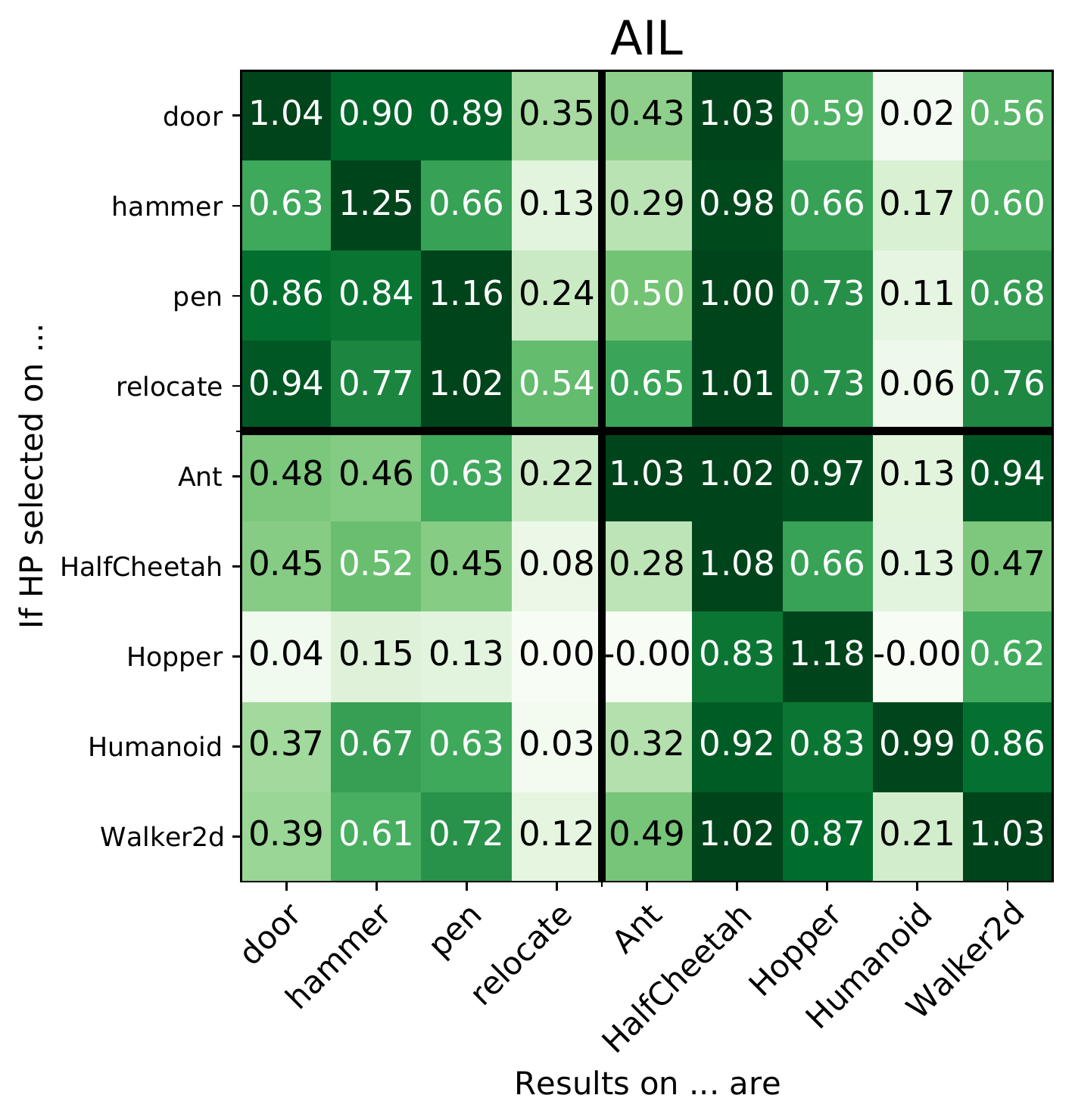}
  \includegraphics[width=0.33\linewidth]{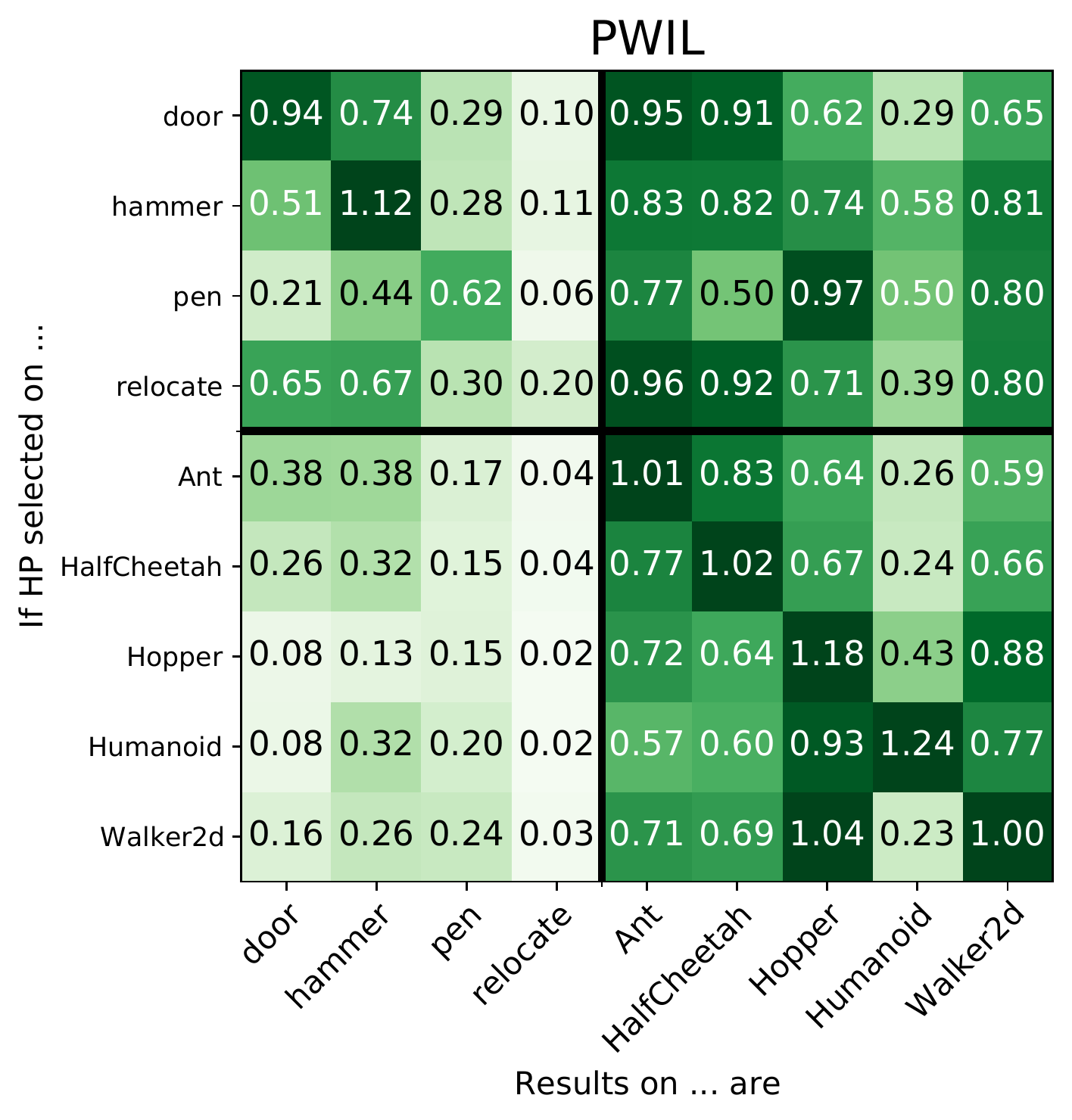}

\end{minipage}%
\caption{HP transfer results for individual validation-test environments pairs.
Rows correspond to different validation environments and columns to different
test environments.
No early stopping is performed.}
\label{fig:transfer_detail_no}
\end{figure*}

We include in~Fig.\ref{fig:transfer-wasserstein} the performance of algorithms when transferring HPs from one environment to another when the early stopping is performed on the state divergence as this metric was the most promising according to Sec.~\ref{subsec:metrics}. We include here the results of the same experiments when early stopping is performed on the oracle return (\ref{fig:transfer_detail_oracle}) or when no early stopping is performed (\ref{fig:transfer_detail_no}).

\end{document}